\documentclass{bmvc2k}

\title{SignRR: Retrieve and Refine Real Motion for Sign Language Production}

\addauthor{Fidel Omar Tito Cruz}{fi513841@ucf.edu}{1}
\addauthor{Angie Sanchez Marquina}{angie.sanchez4@unmsm.edu.pe}{2}
\addauthor{Summy Farfan}{summy.farfan@ucsp.edu.pe}{3}
\addauthor{Gissella Bejarano}{gissella.bejarano@marist.edu}{4}

\addinstitution{
 University of Central Florida\\
 Florida, USA
}
\addinstitution{
 Universidad Nacional Mayor de San Marcos\\
 Lima, Peru
}
\addinstitution{
 Universidad Catolica San Pablo\\
 Arequipa, Peru
}
\addinstitution{
 Marist University\\
 New York, USA
}

\runninghead{TITO CRUZ ET AL.}{RETRIEVE AND REFINE FOR SIGN LANGUAGE PRODUCTION}

\usepackage{booktabs}
\usepackage{multirow}
\usepackage{array}
\usepackage{tabularx}
\usepackage{graphicx}
\usepackage{pifont}
\usepackage{amsmath}
\usepackage{amsfonts}
\usepackage{amssymb}
\usepackage{algorithm}
\usepackage{algpseudocode}
\DeclareMathOperator*{\argmin}{arg\,min}

\newcommand{\cmark}{\ding{51}}
\newcommand{\xmark}{\ding{55}}
\newcommand{\rotcat}[1]{\rotatebox[origin=c]{90}{\textbf{#1}}}

\begin{document}

\maketitle

\begin{abstract}
Sign language production (SLP) aims to generate continuous signing motion from spoken language, often through gloss-to-pose generation. Prior work mainly follows two paradigms. Generative models synthesize motion from a learned prior or from noise, without reference to an observed signing instance, making rare hand configurations and signer-specific articulation difficult to preserve. Retrieval-based methods reuse real, well-articulated motion segments, but concatenating segments from different signers and co-articulation contexts can introduce rhythm and style inconsistencies across the full sequence, not only at segment boundaries. These limitations suggest a complementary solution: use retrieval to provide realistic articulation, and use learned refinement to impose the global coherence that retrieval alone lacks. We therefore propose retrieve-and-refine, a paradigm that starts from real retrieved motion and refines it into a globally coherent signing sequence rather than generating motion from scratch. Our framework, SignRR, initializes motion from a dictionary of real sign segments and refines the full sequence with a part-aware Residual VQ-VAE, where residual quantization preserves fine hand articulation and temporal length differences are handled in the latent space. Experiments on PHOENIX14T and CSL-Daily show that SignRR achieves state-of-the-art back-translation performance while maintaining competitive pose quality.
\end{abstract}

\section{Introduction}
\label{sec:intro}

\begin{figure}[t]
    \centering
    \includegraphics[width=\linewidth]{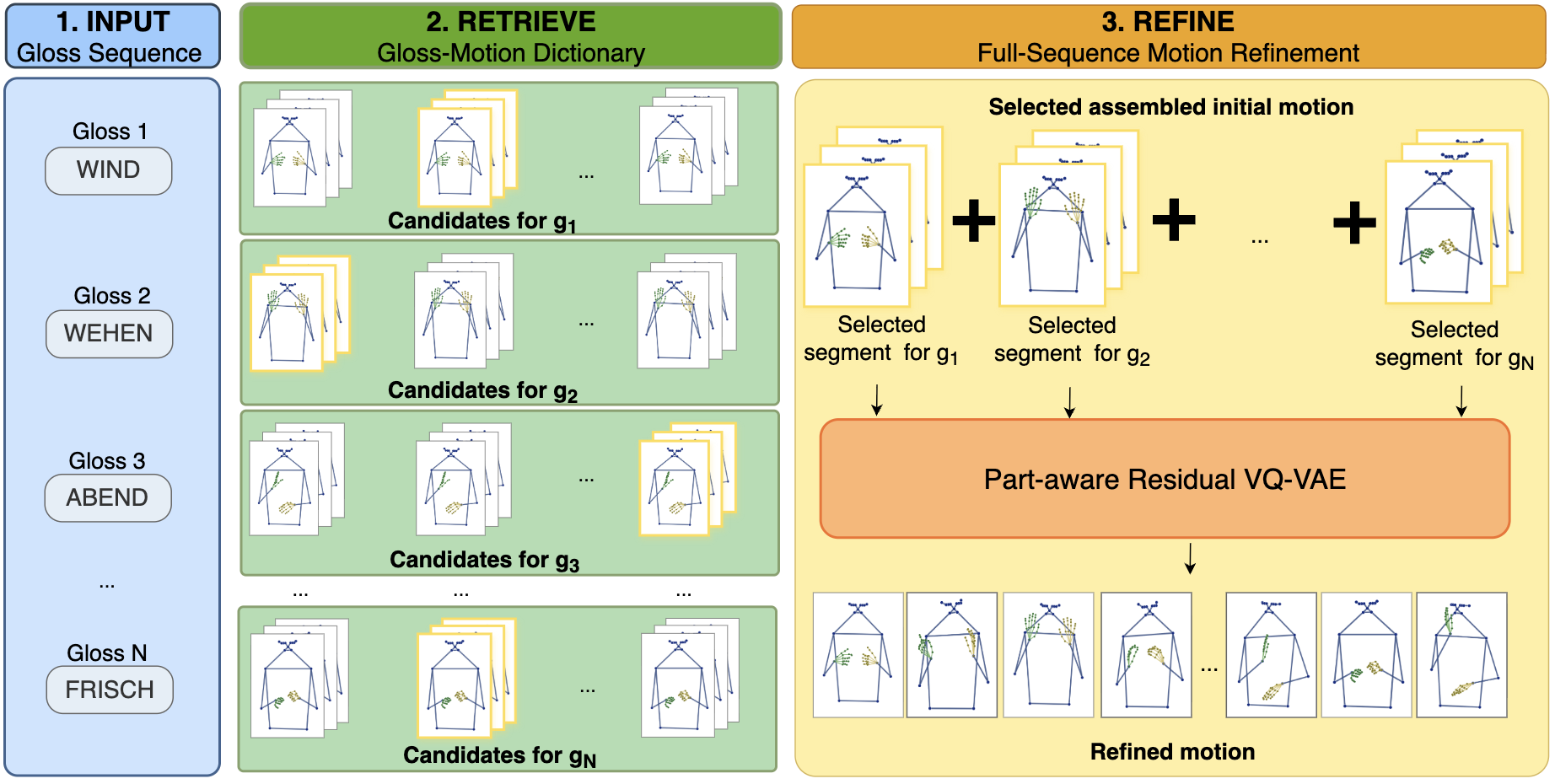}
    \caption{%
    Overview of SignRR. Given an input gloss sequence, SignRR retrieves candidate real
    motion segments from a dictionary, assembles a coherent initial motion, and refines
    it with a part-aware Residual VQ-VAE into the final signing sequence.%
    }
\label{fig:overview}
\end{figure}

Sign language is a crucial means of communication for Deaf communities~\cite{bragg2019sign}. Accordingly, extensive prior work has focused on sign language recognition (SLR) ~\cite{koller2015continuous,zhou2020spatial,cihan2017subunets}, sign language translation (SLT) ~\cite{camgoz2020sign,yin2021including}, and sign language production (SLP) ~\cite{saunders2020progressive,saunders2021mixed}. The latter aims to generate continuous signing motion from spoken language and is commonly formulated as either a text-to-pose (T2P) task, which maps text directly to a pose sequence, or gloss-to-pose task (G2P), which generates pose sequences from intermediate glosses ~\cite{saunders2020progressive}. Because constituent order in signed languages can differ from spoken languages~\cite{sandler2006sign}, directly mapping words to signs is challenging. Glosses, written labels that preserve the sequential structure of signing, reduce this cross-lingual alignment problem to a more tractable pose generation task. Gloss-to-pose has therefore become a widely adopted formulation for SLP and is the setting of this work.\\
 
Sign production from a gloss sequence has been approached in two main paradigms. First, generative methods learn to synthesize the motion directly from the input. Early work regresses pose sequences from glosses ~\cite{saunders2020progressive,saunders2021continuous}, but reconstruction losses drive outputs toward temporally averaged, under-articulated motion, especially in the hands, which are among the most informative cues in sign language ~\cite{sandler2006sign}. Discretizing motion into codebook tokens and treating generation as sequence prediction reduces this averaging effect ~\cite{xie2024g2p,yin2024t2s,zuo2025signs,lee2025glos}, and diffusion-based methods ~\cite{tang2025gloss,tang2025sign} further improve quality. However, this motion is produced from a learned prior or from noise, with no reference to any observed signing instance, so rare hand configurations and signer-specific articulation are compressed into the dominant distribution. Second, retrieval-based methods reuse real sign segments obtained from a dictionary built from training data. Such segments are anatomically plausible, retain signer-specific articulation, and remain semantically faithful. As a result, retrieval-based assembly has been shown to outperform generative methods ~\cite{saunders2022signing,Walsh_2024_BMVC}; the winning submission of the SLRTP 2025 challenge ~\cite{walsh2025slrtp2025} likewise surpassed all generative competitors on back-translation metrics. However, retrieval alone is not enough. Simply concatenating segments from different co-articulation contexts does not guarantee coherent signing and can lead to over-articulated motion. These two paradigms have complementary strengths  (Table~\ref{tab:method_comparison}): the real motion that generation struggles to produce is precisely what retrieval supplies, while the coherent assembly that retrieval cannot guarantee is what a learned model can provide. We build on this complementarity and propose a retrieve-and-refine paradigm, in which the model is asked not to generate signing motion from scratch, but to refine motion that is already real. Retrieved segments provide well-articulated, anatomically valid motion as a starting point, and the learned model is responsible only for resolving the incoherence introduced during assembly, so it never has to synthesize articulation from a prior.\\
 
Turning this idea into an effective method raises three requirements. The first concerns the scope at which refinement must act. Prior work on assembling motion handles boundaries locally, by masking and reconstructing the frames between adjacent segments ~\cite{tang2025discrete,cruz2024generative}. However, when segments come from different signers or are produced at different speeds, the resulting inconsistencies are not confined to the junctions but appear as rhythm mismatches across the full sequence. Refinement must therefore reason over the complete assembled sequence at once, treating the retrieved motion as a structured initialization to be made coherent rather than a set of transitions to be patched. The second requirement concerns the representation in which this refinement is carried out. Signs combine body parts with markedly different kinematics: arm movements unfold over broad temporal windows, whereas finger articulations are rapid and carry lexical distinctions~\cite{sandler2006sign}. Encoding all parts into a shared codebook ~\cite{yin2024t2s} forces hand and body features to compete for capacity, often suppressing the hand details. Even methods that separate body parts ~\cite{xie2024g2p,zuo2025signs,lee2025glos} rely on single-layer quantization, whose reconstruction error is harmful for the hands, where small displacements can change handshape identity. To refine the motion effectively, the representation must be part-aware and precise to preserve fine-grained hand detail. Residual vector quantization provides this precision by encoding the remaining approximation error at each layer ~\cite{zeghidour2021soundstream}. The third requirement is temporal: retrieved segments retain the durations of their source instances, so their concatenation may not match the target sequence length. Refinement must therefore reconcile this duration mismatch as part of making the assembled motion coherent. We design SignRR to meet these requirements. It initializes candidate motion from a constructed gloss-motion dictionary, selects an initial assembly with a lightweight dynamic-programming step, and refines the full sequence with a part-aware Residual VQ-VAE that operates in its quantized latent space, thereby realizing retrieve-and-refine for sign language production (Fig.~\ref{fig:overview}). Our contributions are as follows:
 
\begin{itemize}
 
\item We introduce retrieve-and-refine, a paradigm for sign language production that refines real retrieved motion into coherent signing rather than generating motion from a learned prior, combining the natural articulation of retrieval with the coherence of generation.
 
\item We propose SignRR, a framework that implements this paradigm by initializing motion from a gloss-motion dictionary and refining the full assembled sequence with a part-aware Residual VQ-VAE.
 
\item We present state-of-the-art results on PHOENIX14T and CSL-Daily datasets across standard SLP metrics, with ablation studies validating each component.
 
\end{itemize}
 
\section{Related Work}

\textbf{Sign Language Production. } Early SLP systems relied on avatar-based animation requiring manually specified sign lexicons ~\cite{elliott2008linguistic,mcdonald2016automated}, before data-driven approaches framed gloss-to-pose generation as sequence regression. Saunders et al.~\cite{saunders2020progressive} proposed Progressive Transformers with autoregressive decoding, later extended with a Mixture of Motion Primitives ~\cite{saunders2021mixed} and mixture density networks ~\cite{saunders2021continuous}, yet L2 regression losses consistently drive outputs toward temporally averaged poses, producing under-articulated hands that are the main articulators of lexical meaning in signed languages~\cite{sandler2006sign}. To overcome this, subsequent works discretize motion via vector quantization. G2P-DDM~\cite{xie2024g2p}, trains single-layer VQ-VAEs per body part and generates token sequences with discrete diffusion, T2S-GPT~\cite{yin2024t2s} employs a dynamic-length unified codebook with an autoregressive GPT, SOKE ~\cite{zuo2025signs} tokenizes body and hands separately for multilingual generation, drawing on a sign dictionary only as auxiliary conditioning for its autoregressive generator, and GLOS~\cite{lee2025glos} extends this to four-part streams with temporal-alignment conditioning. Diffusion-based methods ~\cite{tang2025gloss,tang2025sign} synthesize motion from noise under semantic guidance, improving quality in continuous space but requiring many denoising steps and producing output without reference to any signing instance. Across all generative paradigms, infrequent hand configurations and signer-specific motion patterns remain difficult to recover from a learned prior alone.\\

\begin{table}[t]
\centering
\footnotesize
\setlength{\tabcolsep}{3.0pt}
\renewcommand{\arraystretch}{1.05}
\resizebox{\textwidth}{!}{%
\begin{tabular}{c l c c c c l l}
\toprule
 & \textbf{Method} &
\shortstack{\textbf{Real}\\\textbf{art.}} &
\shortstack{\textbf{Signer}\\\textbf{dyn.}} &
\shortstack{\textbf{Boundary}\\\textbf{smooth.}} &
\shortstack{\textbf{Global}\\\textbf{coh.}} &
\textbf{Approach} &
\textbf{Coherence handling} \\
\midrule

\multirow{4}{*}{\rotcat{Generative}}
& Progressive Transf.~\cite{saunders2020progressive}
& \xmark & \xmark & \cmark & \cmark
& Autoregressive regression
& Generated jointly \\

& G2P-DDM~\cite{xie2024g2p}
& \xmark & \xmark & \cmark & \cmark
& VQ + discrete diffusion
& Generated jointly \\

& SOKE~\cite{zuo2025signs}
& \xmark & \xmark & \cmark & \cmark
& Body/hand VQ and AR gen.
& Generated jointly \\

& Sign-IDD~\cite{tang2025sign}
& \xmark & \xmark & \cmark & \cmark
& Diffusion from noise
& Generated jointly \\

\midrule

\multirow{4}{*}{\rotcat{Retrieval}}
& Saunders~\cite{saunders2022signing}
& \cmark & \cmark & \xmark & \xmark
& Neural frame selection
& Interpolation \\

& Sign Stitching~\cite{Walsh_2024_BMVC}
& \cmark & \cmark & \xmark & \xmark
& Signal-processing stitching
& Local smoothing \\

& SLRTP'25 winner~\cite{walsh2025slrtp2025}
& \cmark & \cmark & \xmark & \xmark
& CTC-dictionary and stitching
& Local stitching \\

& Gloss-driven~\cite{tang2025gloss}
& \cmark & \cmark & \cmark & \xmark
& Retrieval + boundary diffusion
& Per-junction refinement \\

\midrule

\raisebox{-0.45ex}{\rotcat{Ours}}
& \rule{0pt}{3.2ex}SignRR
& \cmark & \cmark & \cmark & \cmark
& DP assembly and RVQ-VAE
& Full-sequence refinement \\

\bottomrule
\end{tabular}%
}
\caption{
Comparison of sign language production paradigms.
Real art. denotes preservation of real signing articulation; signer dyn. denotes preservation of signer-specific dynamics; boundary smooth. denotes explicit boundary smoothness between motion units; and global coh. denotes sequence-level coherence beyond local transitions.
}
\label{tab:method_comparison}
\end{table}
 
\textbf{Retrieval-based Sign Production.}
An alternative to generating motion from a learned prior is to assemble signing motion from real segments stored in a gloss-motion dictionary. Stoll et al.~\cite{stoll2020text2sign} introduced a motion-graph formulation in which dynamic programming selects an optimal sequence of sign segments. Saunders et al. ~\cite{saunders2022signing} scaled this idea with a frame selection network supervised by DTW alignment over an interpolated dictionary of signs, while Walsh et al.~\cite{Walsh_2024_BMVC} replaced neural selection with a signal-processing pipeline based on canonical normalization, velocity-bounded interpolation, and frequency-domain filtering. Dictionaries can also be built from continuous signing: Walsh et al.~\cite{walsh2024data} learn a discrete motion vocabulary with NSVQ, and Zuo et al.~\cite{zuo2024towards} derive pseudo-temporal boundaries from a CTC recognition model, a strategy later adopted by the SLRTP 2025 challenge winner ~\cite{walsh2025slrtp2025}. However, existing retrieval-based methods mainly focus on selecting segments and stitching them with local or hand-crafted transitions, such as linear interpolation, spline fitting, or frequency-domain smoothing. This preserves real articulation, but does not ensure coherence across the complete sequence, especially when segments come from different signers, speeds, or co-articulation contexts. Tang et al.~\cite{tang2025discrete} learn boundary transitions with conditional diffusion, but still operate only on individual junctions rather than on the assembled sequence as a whole. SignRR addresses this limitation by treating retrieval as an initialization and refining the complete assembled motion with a learned model, targeting sequence-level rhythm and articulation inconsistencies that local stitching cannot resolve.\\

\textbf{Motion Representations for SLP.} The general motion generation literature offers architectural precedents for discrete motion modelling. Van den Oord et al.~\cite{van2017neural} established vector quantization with straight-through gradient estimation, and Zeghidour et al.~\cite{zeghidour2021soundstream} extended this with residual vector quantization, where cascaded codebooks encode successive residuals for progressive fidelity without codebook explosion. Guo et al.~\cite{guo2024momask} show that six-layer residual VQ substantially outperforms single-layer VQ for motion reconstruction, Lu et al.~\cite{lu2023humantomato} demonstrate that separating body and hand codebooks improves whole-body generation, and Zou et al. ~\cite{zou2024parco} show that coordinating separate body-part codebooks improves overall limb coherence. These models, however, assume generation from a clean learned prior, an assumption that does not hold when the input is a retrieved sequence assembled from segments of different signers and co-articulation contexts. In SLP, G2P-DDM~\cite{xie2024g2p}, SOKE~\cite{zuo2025signs}, and GLOS~\cite{lee2025glos} use single-layer per-part quantization, while residual quantization has appeared in sign only for pose interpolation~\cite{cruz2024generative}, a different task from gloss-to-pose production. SignRR instead uses part-aware residual quantization to refine real assembled motion, treating retrieval as a structured initialization rather than noise to be denoised.\\

\textbf{Pose-Conditioned Human Generation.} Generated pose can also be used as a conditioning signal for human appearance generation. In sign language, Stoll et al.~\cite{stoll2020text2sign} render signer videos from poses with a conditional GAN. In addition, IMAGPose~\cite{shen2024imagpose} conditions person image generation on target poses, while IMAGDressing-v1~\cite{shen2025imagdressing} supports pose as one of several controls for editable human synthesis. SignRR focuses on generating the pose sequence itself, which can serve as a structured motion representation for such downstream applications.

\section{Method}\label{sec:method}
 
Given an input gloss sequence $G = (g_1, \dots, g_N)$, our goal is to produce a pose sequence $\hat{\mathbf{x}} = (\hat{x}_1, \dots, \hat{x}_{\hat{T}})$, where $\hat{x}_t \in \mathbb{R}^{K \times 3}$ and $K{=}61$. SignRR follows a retrieve-and-refine pipeline (Fig.~\ref{fig:overview}), where real sign segments are first retrieved from a gloss-motion dictionary and assembled into an initial motion sequence, which a learned refinement model then processes as a whole to resolve boundary discontinuities, rhythm mismatches, and inconsistencies across body parts. The method has three steps: an offline gloss-motion dictionary $\mathcal{D}$ built from continuous training data (Sec.~\ref{sec:dictionary}); a part-aware Residual VQ-VAE trained to refine assembled motion into target motion (Sec.~\ref{sec:refine}); and an inference procedure that selects a coherent assembly by dynamic programming before refinement (Sec.~\ref{sec:inference}). We write $\mathbf{x}_a$ and $\mathbf{x}_b$ for the assembled and target motion sequences, of length $T_a$ and $T_b$, and let $D{=}K{\times}3{=}183$ be the flattened pose dimension.

\subsection{Gloss-Motion Dictionary}\label{sec:dictionary}
 
We construct $\mathcal{D}$ automatically from the continuous signing sequences in the training set, so that the extracted segments retain co-articulated motion rather than isolated sign realizations, where each sign is produced on its own and lacks the natural transitions of continuous signing. Segment boundaries come from a pretrained CorrNet+~\cite{hu2024corrnet+} CSLR recognizer: given the ground-truth gloss sequence, we mark the start of each gloss at the first frame where the recognizer's logit for that gloss exceeds the logit of the previous one. Because this is a ground-truth-constrained alignment rather than CTC decoding, it recovers segment timing from the recognizer's logits alone and needs only the sentence-level gloss sequences already provided by the dataset, not frame-level boundary annotations. For each gloss, we then score its candidate segments by motion smoothness and duration regularity, discard outliers, and retain $S$ diverse, high-quality representatives ($|\mathcal{D}(g)|{=}S$), caching a short boundary descriptor of each for the inference-time selector. The segmentation rule, quality score, and curation procedure are detailed in the supplementary.

\subsection{Part-aware Residual VQ-VAE}\label{sec:refine}

\begin{figure}[t]
    \centering
    \includegraphics[width=\linewidth]{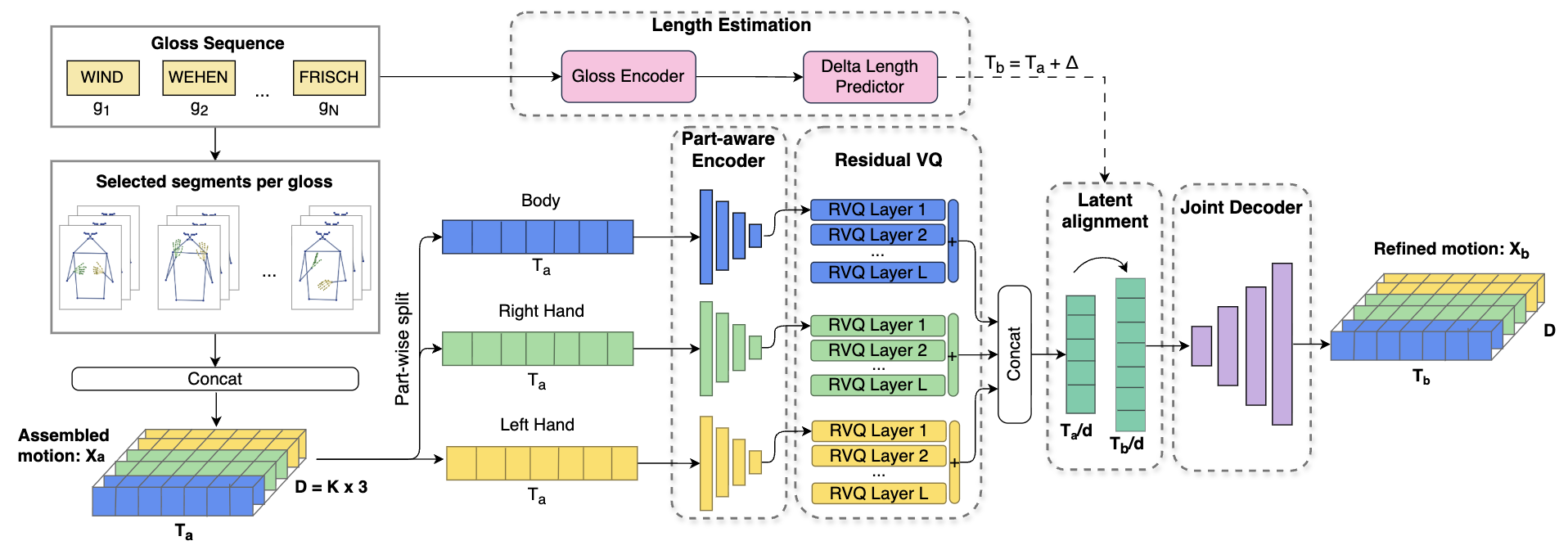}
    \caption{%
    Architecture of SignRR's refinement model, a part-aware Residual VQ-VAE. The assembled motion is split into body, right-hand, and left-hand streams each encoded and discretized by its own residual VQ. In parallel, a gloss encoder estimates the target length; the per-part latents are aligned to this length and decoded by a single joint decoder into the refined signing sequence.%
    }
    \label{fig:arch}
\end{figure}

\smallskip\noindent\textbf{Part-aware hierarchical encoder.}
We encode the assembled motion with separate streams for the body and hands (Fig.~\ref{fig:arch}). This design reflects the different roles of these parts in signing: body and arm motion provide the main pose trajectory, while fine-grained changes in handshape, location, and movement can yield lexical contrasts in sign language~\cite{sandler2006sign}. Encoding the whole pose with a single codebook can make these fine hand patterns harder to preserve, since they are modeled together with larger body movements. We therefore split the input $\mathbf{x} \in \mathbb{R}^{T \times D}$ along the feature axis into body ($D_{\mathrm{body}}{=}57$), right hand ($D_{\mathrm{rh}}{=}63$), and left hand ($D_{\mathrm{lh}}{=}63$), and pass each part through a dedicated encoder $Enc^p$:

\begin{equation}\label{eq:encoder}
    \mathbf{z}^p = Enc^p(\mathbf{x}^p) \;\in\; \mathbb{R}^{\frac{T}{d} \times d_p}\;,
\quad p \in \{\text{body},\, \text{rh},\, \text{lh}\}\;,
\end{equation}
where $d_p$ is the latent dimension of part $p$ and $d{=}4$ is the temporal downsampling factor. Each encoder applies an initial 1D convolution, two stride-2 downsampling stages with dilated ResNet-1D blocks, and a projection to $d_p$. The body encoder uses wider dilation rates to cover longer-range arm and torso motion, while the hand encoders use narrower dilation rates for faster hand motion. The hand streams are also assigned larger latent dimensions and codebooks.

\smallskip\noindent\textbf{Residual vector quantization.}
For each part latent $\mathbf{z}^p$, we use residual vector quantization with $L$ cascaded codebooks $C^p_l \subset \mathbb{R}^{d_p}$. Instead of representing the latent with a single codebook, successive layers encode the remaining residual, which helps preserve fine pose details~\cite{zeghidour2021soundstream,guo2024momask}:

\begin{equation}\label{eq:rvq}
    \hat{\mathbf{z}}^p_l = \mathrm{VQ}\!\bigl(r_{l-1}^p;\, C^p_l\bigr)\;, \qquad
    r_l^p = r_{l-1}^p - \mathrm{sg}\!\bigl[\hat{\mathbf{z}}^p_l\bigr]\;, \qquad l = 1,\dots,L\;,
\end{equation}
with $r_0^p = \mathbf{z}^p$ denoting the initial residual. The operator $\mathrm{VQ}(\cdot\,;C)$ maps its input to the nearest entry in $C$ under $\ell_2$ distance and is trained with the straight-through estimator, while $\mathrm{sg}[\cdot]$ denotes stop-gradient. The full latent is obtained by summing the quantized outputs across layers for each part and concatenating the resulting part-wise latents:

\begin{equation}\label{eq:concat}
    \hat{\mathbf{z}}^p = \sum_{l=1}^{L} \hat{\mathbf{z}}^p_l\;, \qquad
    \hat{\mathbf{z}} = \bigl[\hat{\mathbf{z}}^{\mathrm{body}};\, \hat{\mathbf{z}}^{\mathrm{rh}};\, \hat{\mathbf{z}}^{\mathrm{lh}}\bigr]
    \in \mathbb{R}^{C \times T_a/d}\;, \qquad C = \textstyle\sum_{p} d_p\;.
\end{equation}
 
\smallskip\noindent\textbf{Latent alignment.}
The assembled input and the target generally differ in length: $\mathbf{x}_a$ is concatenated from segments retrieved independently per gloss, each contributed at its own duration, so its total length need not match that of $\mathbf{x}_b$. We reconcile the two in the latent space in two steps. First, a lightweight Transformer gloss encoder maps $G$ to contextual features whose mean over the sequence, $\bar{\mathbf{h}}$, feeds an MLP that predicts the length difference $\Delta$ (the Length Estimation
block in Fig.~\ref{fig:arch}):
\begin{equation}\label{eq:delta}
    \Delta = \mathrm{MLP}\!\bigl([\bar{\mathbf{h}};\; T_a / T_{\max}]\bigr) \cdot \delta_{\mathrm{scale}}\;,
\end{equation}
where $T_{\max}$ is a fixed length normalizer and $\delta_{\mathrm{scale}}$ bounds the range. Second, the quantized latent is resampled to the target temporal resolution by linear interpolation with aligned endpoints (the Latent alignment block in Fig.~\ref{fig:arch}):
\begin{equation}\label{eq:interp}
    \tilde{\mathbf{z}} = \mathrm{Interpolate}\!\bigl(\hat{\mathbf{z}},\; T_b/d\bigr) \;\in\; \mathbb{R}^{C \times T_b/d}\;.
\end{equation}
During training the target length $T_b$ is known and used directly; at inference it is unknown and replaced by the predicted length $\hat{T}_b = T_a + \Delta$. Applying the adjustment to the quantized latent rather than to raw joint coordinates lets it operate on compact motion codes, while the RVQ bottleneck regularizes the representation before resampling. Gloss information is therefore used only to predict the length adjustment, while the decoder operates solely on the aligned motion latent.
 
\smallskip\noindent\textbf{Joint decoder.} 
A single decoder maps the aligned latent back to motion, so that coordination across parts is recovered jointly rather than per stream:
\begin{equation}\label{eq:decoder}
    \hat{\mathbf{x}} = \mathrm{Dec}(\tilde{\mathbf{z}}) \in \mathbb{R}^{T_{\mathrm{out}} \times D},
\end{equation}
 
where $T_{\mathrm{out}}=T_b$ during training and $T_{\mathrm{out}}=\hat{T}_b$ at inference.
The decoder applies an initial projection, two upsampling stages with dilated ResNet-1D blocks that mirror
the encoder, and a final projection to the $D{=}183$ pose dimensions.
 
\smallskip\noindent\textbf{Training objective.} For each training sample, we draw one segment per gloss from $\mathcal{D}$ and concatenate them to form the assembled input $\mathbf{x}_a$, using the corresponding real continuous sequence $\mathbf{x}_b$ as supervision. This trains the model to map different retrieved assemblies to coherent target motion while retaining retrieval as the articulated starting point. Resampling across epochs exposes the model to various combinations of segments and assembly patterns. We minimize
\begin{equation}\label{eq:loss}
    \mathcal{L} = \mathcal{L}_{\mathrm{rec}} + \lambda_v \mathcal{L}_{\mathrm{vel}} + \lambda_c \mathcal{L}_{\mathrm{commit}} + \lambda_\ell \mathcal{L}_{\mathrm{length}}\;,
\end{equation}
where $\mathcal{L}_{\mathrm{rec}}$ is a part-weighted smooth-$\ell_1$ reconstruction loss that weights the
hands above the body, $\mathcal{L}_{\mathrm{vel}}$ penalizes frame-to-frame velocity error, and
$\mathcal{L}_{\mathrm{commit}}$ is the RVQ commitment term summed over all codebooks. The final term,
$\mathcal{L}_{\mathrm{length}} = \|\Delta - (T_b - T_a)\|_1$, supervises the length predictor; since
training interpolates to the ground-truth length, it updates only the gloss-encoder and predictor branch
and does not perturb the reconstruction path. Architecture sizes, codebook configuration, and loss
weights are given in the supplementary.

\subsection{Inference: Assembly and Refinement}\label{sec:inference}
 
At inference, SignRR first selects a set of dictionary segments by dynamic programming and then refines
the assembled sequence with the trained model.
 
\smallskip\noindent\textbf{Segment selection.}
For a gloss sequence $G$, we cast assembly as a shortest-path problem over a trellis whose nodes at
position $i$ are the candidate segments $\mathcal{D}(g_i)$. The selected path
$\pi^* = (\pi^*_1, \dots, \pi^*_N)$, $\pi_i \in \mathcal{D}(g_i)$, minimizes a combined boundary-smoothness
and quality cost:
\begin{equation}\label{eq:dp_cost}
    \pi^* = \argmin_{\pi} \sum_{i=2}^{N} \left[ \alpha \cdot \frac{d_{\mathrm{bnd}}(\pi_{i-1},\, \pi_i)}{\tilde{d}} + \beta \cdot \bigl(1 - q(\pi_i)\bigr) \right] + \beta \cdot \bigl(1 - q(\pi_1)\bigr)\;,
\end{equation}
where $d_{\mathrm{bnd}}$ is the mean $\ell_2$ distance between the boundary descriptors of adjacent
segments, $q(\cdot)$ the cached quality score, and $\tilde{d}$ a normalizer estimated from $\mathcal{D}$.
The path is found by standard dynamic programming in $O(N S^2)$; out-of-dictionary glosses are handled as
described in the supplementary.
 
\smallskip\noindent\textbf{Refinement.} The selected segments are concatenated into $\mathbf{x}_a$ and passed through the trained model. The model
encodes the sequence with the part-specific encoders, applies residual vector quantization, aligns the
latent sequence to the predicted length, and decodes it into the final pose sequence
$\hat{\mathbf{x}} \in \mathbb{R}^{\hat{T}_b \times D}$. Refining the full assembled sequence
in the quantized latent space, rather than patching individual junctions, is what reduces the rhythm and
articulation inconsistencies introduced during assembly.

\section{Experiments and Results}

\subsection{Experimental Setup}
\noindent\textbf{Datasets.}
We evaluate SignRR on two standard SLP benchmarks: PHOENIX14T~\cite{camgoz2018neural} and CSL-Daily~\cite{zhou2021improving}. PHOENIX14T contains $7{,}096$ training, $519$ validation, and $642$ test sequences over $1{,}085$ glosses, while CSL-Daily contains $18{,}401$, $1{,}077$, and $1{,}176$ sequences, respectively, over $2{,}000$ glosses. Both datasets provide sentence-level gloss annotations. Following prior SLP work, 2D keypoints are extracted with OpenPose~\cite{cao2017realtime} and lifted to 3D using the pose-lifting procedure of Ivashechkin et al.~\cite{ivashechkin2023improving}. Each frame is represented as a $K{=}61$-keypoint 3D pose, consisting of $19$ body joints and $21$ joints for each hand, giving $D{=}183$ pose values. The gloss-motion dictionary $\mathcal{D}$ is built once per dataset from the training split.

\smallskip\noindent\textbf{Evaluation metrics.}
Following the back-translation protocol used in prior SLP work~\cite{camgoz2020sign,tang2025sign}, we train a Sign Language Transformer on real poses and evaluate it on the generated poses, comparing the recovered text with the reference. We report WER, BLEU-1, BLEU-4, ROUGE-L, and MPJPE.

\smallskip\noindent\textbf{Implementation details.}
For dictionary construction, we use the publicly released CorrNet+~\cite{hu2024corrnet+} checkpoints without further fine-tuning. For our refinement model, we use three part-specific encoder streams for the body, right hand, and left hand, with latent widths $d_{\mathrm{body}}{=}192$ and $d_{\mathrm{rh}}{=}d_{\mathrm{lh}}{=}256$. The body stream uses dilation rates $[1,3,9]$ to capture long-range arm and torso motion, while the hand streams use $[1,3,5]$ for faster finger articulation. Each part is quantized with $L{=}3$ residual codebooks. We train with AdamW and batch size $256$. We trained for $800$ epochs on PHOENIX14T and $400$ epochs on CSL-Daily on a single NVIDIA A100 GPU. Additional architecture and optimization details are provided in the supplementary.

\subsection{Comparison with State-of-the-Arts}

\begin{table*}[!t]
\centering
\caption{Comparison with state-of-the-art SLP methods on PHOENIX14T. $\uparrow$ and $\downarrow$ indicate higher and lower is better, respectively. 
\textbf{Bold} and \underline{underlined} values denote the best and second-best results among non-GT methods with reported values. 
Methods marked with $\dagger$ are reproduced under our evaluation setting; the remaining baseline results are taken from~\cite{Walsh_2024_BMVC}.}
\label{tab:slp_results}
\resizebox{\textwidth}{!}{%
\begin{tabular}{l|ccccc|ccccc}
\toprule
\multirow{2}{*}{\textbf{Method}}
  & \multicolumn{5}{c|}{\textbf{DEV}}
  & \multicolumn{5}{c}{\textbf{TEST}} \\
\cmidrule(lr){2-6} \cmidrule(lr){7-11}
  & WER$\downarrow$ & BLEU-1$\uparrow$ & BLEU-4$\uparrow$ & ROUGE$\uparrow$ & MPJPE$\downarrow$
  & WER$\downarrow$ & BLEU-1$\uparrow$ & BLEU-4$\uparrow$ & ROUGE$\uparrow$ & MPJPE$\downarrow$ \\
\midrule
GT
  & 64.67 & 30.88 & 13.38 & 31.11 & 0.00
  & 63.61 & 31.30 & 13.43 & 31.03 & 0.00 \\
\midrule
Sign-IDD$^{\dagger}$~\cite{tang2025sign}
  & 86.34 & 26.49 & 10.79 & 27.34 & \textbf{40.74}
  & 85.98 & 25.34 &  9.81 & 24.91 & \textbf{50.19} \\
G2P-DDM$^{\dagger}$~\cite{xie2024g2p}
  & \underline{85.43} & \underline{26.97} & \underline{11.05} & \underline{28.13} & 44.41
  & \underline{83.92} & \underline{27.04} & \underline{10.78} & \underline{27.56} & 54.28 \\
Miao et al.$^{\dagger}$~\cite{miao2026twostage}
  & 95.62 & 15.97 &  4.60 & 16.94 & 45.57
  & 95.66 & 14.16 &  3.29 & 14.61 & 56.17 \\
% \midrule
Sign Stitching~\cite{Walsh_2024_BMVC}
  & -- & -- & -- & -- & --
  & -- & 25.14 &  6.67 & 26.49 & -- \\
PT base~\cite{saunders2020progressive}
  & -- & -- & -- & -- & --
  & -- &  6.27 &  1.59 &  9.50 & -- \\
PT + GN~\cite{saunders2020progressive}
  & -- & -- & -- & -- & --
  & -- & 11.45 &  4.04 & 14.52 & -- \\
\midrule
\textbf{Ours}
  & \textbf{82.07} & \textbf{28.16} & \textbf{11.72} & \textbf{28.88} & \underline{43.83}
  & \textbf{80.98} & \textbf{28.78} & \textbf{10.96} & \textbf{28.15} & \underline{53.70} \\
\bottomrule
\end{tabular}%
}
\end{table*}

\begin{table*}[!t]
\centering
\caption{Comparison with state-of-the-art SLP methods on CSL-Daily. $\uparrow$ and $\downarrow$ indicate higher and lower is better, respectively. \textbf{Bold} and \underline{underlined} values denote the best and second-best results among non-GT methods. Methods marked with $\dagger$ are reproduced under our evaluation setting.}
\label{tab:slp_results_csl}
\resizebox{\textwidth}{!}{%
\begin{tabular}{l|cccc|cccc}
\toprule
\multirow{2}{*}{\textbf{Method}}
  & \multicolumn{4}{c|}{\textbf{DEV}}
  & \multicolumn{4}{c}{\textbf{TEST}} \\
\cmidrule(lr){2-5} \cmidrule(lr){6-9}
  & BLEU-1$\uparrow$ & BLEU-4$\uparrow$ & ROUGE$\uparrow$ & MPJPE$\downarrow$
  & BLEU-1$\uparrow$ & BLEU-4$\uparrow$ & ROUGE$\uparrow$ & MPJPE$\downarrow$ \\
\midrule
GT
  & 23.59 & 5.87 & 23.51 & 0.00
  & 23.12 & 5.06 & 23.06 & 0.00 \\
\midrule
Sign-IDD$^{\dagger}$~\cite{tang2025sign}
  & \underline{11.66} & 0.00 & \underline{11.83} & 135.71
  & \underline{11.98} & 0.00 & \underline{11.98} & 147.51 \\
Miao et al.$^{\dagger}$~\cite{miao2026twostage}
  & 11.43 & \underline{0.23} & 11.30 & \textbf{91.29}
  & 11.53 & \underline{0.20} & 11.46 & \textbf{99.70} \\
\midrule
\textbf{Ours}
  & \textbf{15.23} & \textbf{1.34} & \textbf{15.01} & \underline{106.21}
  & \textbf{15.45} & \textbf{1.57} & \textbf{15.46} & \underline{115.40} \\
\bottomrule
\end{tabular}%
}
\end{table*}

\begin{figure}[!t]
    \centering
    \includegraphics[width=0.95\linewidth]{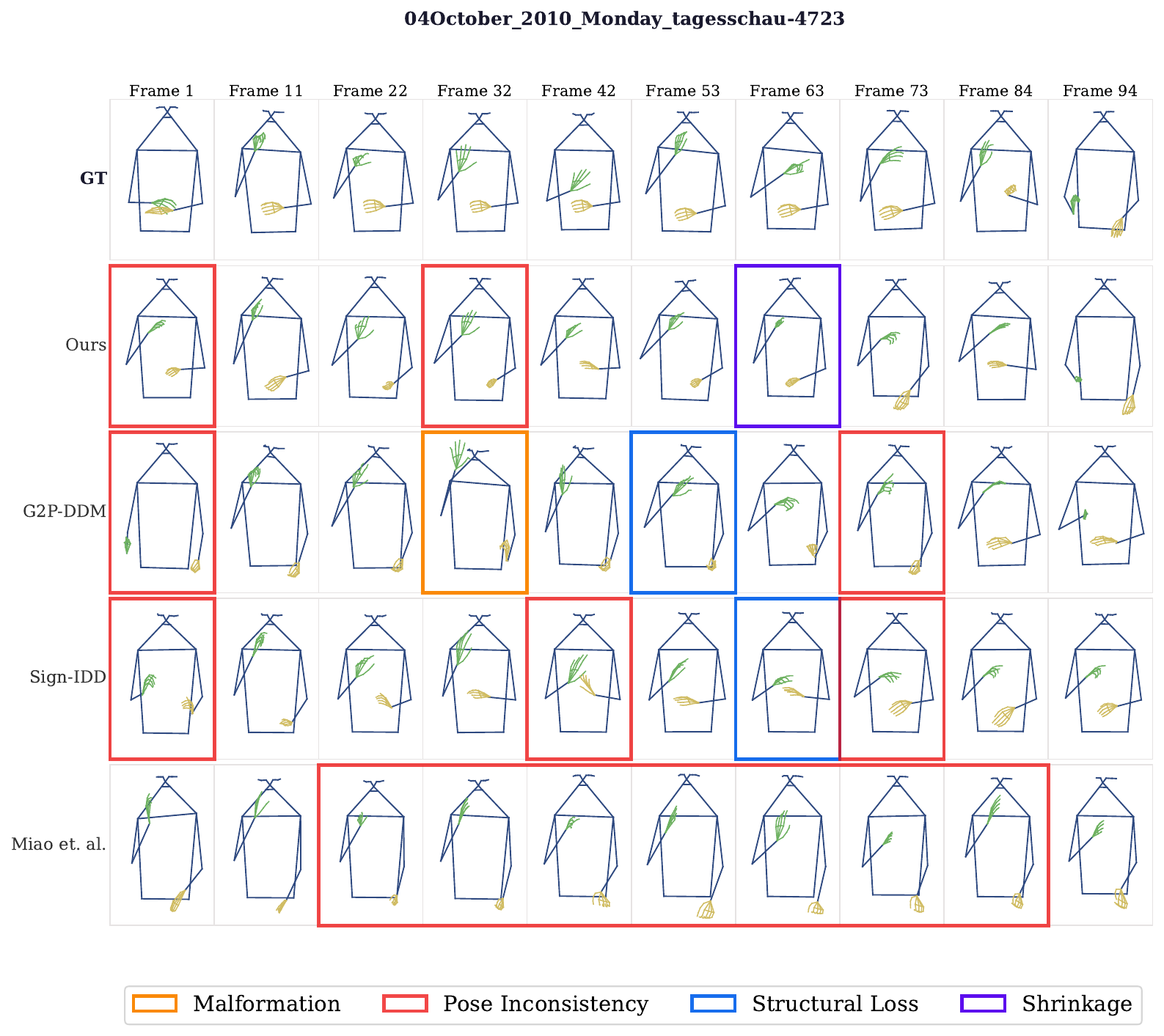}
    \caption{Qualitative comparison of sign language production models on PHOENIX-14T.}
    \label{Qualitative_phoenix}
\end{figure}

\smallskip\noindent\textbf{Quantitative Results.}
Table~\ref{tab:slp_results} compares SignRR with prior SLP methods on PHOENIX14T. All methods are evaluated with the same $K{=}61$-keypoint pose representation, the same 2D-to-3D lifting pipeline, and a shared back-translation model. Methods marked with $\dagger$ are reproduced from released code under this setting, while Sign Stitching and the PT baselines are reported from~\cite{Walsh_2024_BMVC}, which follows the same pose and back-translation protocol. The GT row is included as a reference ceiling for the back-translation protocol.

SignRR achieves the best back-translation performance among all non-GT methods on both splits. On the test set, it reaches WER $80.98$, BLEU-1 $28.78$, BLEU-4 $10.96$, and ROUGE $28.15$, improving over G2P-DDM across all four metrics. The gain is especially clear over the regression-based PT baselines, whose low BLEU and ROUGE scores suggest that pose smoothness alone does not preserve sign content. Compared with the retrieval-based Sign Stitching baseline, SignRR improves test BLEU-4 from $6.67$ to $10.96$, suggesting that full-sequence refinement is more effective than local stitching alone. Sign-IDD obtains the lowest MPJPE among non-GT methods, but SignRR outperforms it on all back-translation metrics. Overall, SignRR ranks second on MPJPE while achieving the strongest back-translation scores, showing a better balance between pose accuracy and intelligibility.

On CSL-Daily (Table~\ref{tab:slp_results_csl}), we report the same metrics, except WER. SignRR again achieves the best back-translation performance among all non-GT methods on both splits. On the test set, it reaches BLEU-1 $15.45$, BLEU-4 $1.57$, and ROUGE $15.46$, improving over both Sign-IDD and Miao et al.\ on every back-translation metric. BLEU-4 values are low because the back-translation ceiling on CSL-Daily is also low, with GT reaching only $5.06$ on the test set. Even under this limited ceiling, SignRR obtains the highest BLEU-4 among non-GT methods and also leads on BLEU-1 and ROUGE. As on PHOENIX14T, the lowest MPJPE does not lead to the best back-translation performance: Miao et al.\ obtains the lowest MPJPE among non-GT methods, while SignRR ranks second and achieves the strongest back-translation scores. This shows the same overall trend on the larger and more lexically diverse CSL-Daily dataset.

\smallskip\noindent\textbf{Qualitative Results.}
Figure~\ref{Qualitative_phoenix} shows that SignRR preserves hand and upper-body structure more reliably than the compared methods across the sampled frames. The colored annotations mark four common error types, including malformation, pose inconsistency, structural loss, and shrinkage. SignRR exhibits fewer of these errors, with mostly mild local issues and no clear malformation or structural loss. By contrast, G2P-DDM shows the widest range of failures, including malformed hands, structural loss, and pose inconsistencies. Sign-IDD and Miao et al.\ avoid clear malformation in this example, but still show repeated pose inconsistencies. In Miao et al., these inconsistencies persist across a longer part of the sequence, with hand poses that change little over time.

In addition to the failures highlighted above, the videos show a small temporal offset from the ground truth across the compared methods. The generated signs often appear a few frames behind the reference, suggesting that temporal alignment remains a shared challenge for current models. Overall, SignRR does not exactly match the ground truth, but it avoids many hand and pose failures observed in the baselines.

% Figure \ref{Qualitative_phoenix} presents qualitative results for SignRR, G2P-DDM, Sign-IDD, and Two-stage on the PHOENIX-14T test set. The analysis focuses on four aspects: hand morphology, motion fluency, temporal alignment, and fidelity to the ground truth.

% Regarding hand morphology, SignRR preserves hand configurations most consistently, with only occasional shrinkage artifacts where the hand appears smaller than expected. Sign-IDD retains hand shape reasonably well but sometimes loses structural detail, with fingers that are not correctly positioned. G2P-DDM shows the most severe distortions, including broken or unnaturally bent fingers across multiple frames. Two-stage produces frequent artifacts, often failing to form recognizable hand shapes altogether.

% In terms of motion fluency, SignRR produces smooth, natural transitions between signs that closely follow the ground truth. Sign-IDD introduces skeletal jitter, small, unsteady movements that disrupt the flow of signing. Two-stage struggles to reproduce entire sequences coherently, often generating incorrect or static hand positions. GSP-DDM produces stiff, mechanical movements that lack the fluid, continuous quality characteristic of natural signing.

% Regarding temporal alignment, a consistent offset of a few frames is observed across all four models, where the generated signs appear slightly behind the ground truth. This suggests a shared limitation of current approaches rather than a model-specific issue.

\subsection{Ablation Studies}

Table~\ref{tab:ablation} studies the refinement model on PHOENIX14T by varying four components: the multi-part encoder (MP), the number of residual codebooks $N_q$, the encoder dilation setting, and the inference-time length source. All variants use the same gloss-motion dictionary and dynamic-programming assembly.

\smallskip\noindent\textbf{Multi-part encoder.} Replacing the separate body, right-hand, and left-hand streams with a single encoder over the full pose reduces BLEU-1 from $28.78$ to $27.89$ and ROUGE from $28.15$ to $26.94$, while MPJPE remains nearly unchanged ($53.70$ and $53.52$). This suggests that a shared encoder can still recover the average pose geometry, but is less effective at preserving the hand details that are important for lexical recognition. The part-aware design therefore improves back-translation without relying on lower joint error alone.

\smallskip\noindent\textbf{Number of residual codebooks.} Using a single codebook ($N_q{=}1$) gives the weakest back-translation results, with BLEU-1 $26.82$ and ROUGE $26.58$. Increasing the number of codebooks improves the representation: $N_q{=}2$ already approaches the proposed setting and obtains the best BLEU-4 ($11.23$), while $N_q{=}3$ gives the strongest BLEU-1 ($28.78$), CHRF ($32.90$), and ROUGE ($28.15$). Increasing to $N_q{=}4$ slightly improves WER, but lowers BLEU-1, BLEU-4, CHRF, and ROUGE. We therefore use $N_q{=}3$, which provides the best overall trade-off across back-translation metrics.

\smallskip\noindent\textbf{Encoder dilation.} The proposed model uses a mixed dilation setting, with $[1,3,9]$ for the body stream and $[1,3,5]$ for the hand streams. Using $[1,3,9]$ for all streams lowers BLEU-1 to $27.63$ and ROUGE to $27.13$, while using $[1,3,5]$ for all streams gives BLEU-1 $27.90$ and ROUGE $27.85$. The mixed setting performs best on BLEU-1, CHRF, and ROUGE, supporting the use of different temporal receptive fields for body and hand motion.

\smallskip\noindent\textbf{Length prediction.} Replacing the predicted length with the ground-truth length improves pose-quality metrics, reducing MPJPE from $53.70$ to $47.61$ and FID from $1.21$ to $1.16$. However, it lowers back-translation performance, with BLEU-1 decreasing from $28.78$ to $27.07$, BLEU-4 from $10.96$ to $10.28$, and ROUGE from $28.15$ to $26.68$. This shows that better frame-level alignment to the reference does not necessarily improve sign intelligibility under the back-translation protocol. A likely explanation is that the predicted length yields temporal dynamics that better support recognition, even when it increases frame-level pose error.

\begin{table*}[!t]
\centering
\caption{Ablation study of SignRR on PHOENIX14T (\textsc{Test}). 
MP denotes the multi-part encoder, $N_q$ the number of RVQ codebooks, Dilation the encoder dilation setting, and Length the inference-time length source.}
\label{tab:ablation}
\resizebox{\textwidth}{!}{%
\begin{tabular}{l|cccc|ccccc|ccc}
\toprule
\multirow{2}{*}{\textbf{Variant}}
  & \multicolumn{4}{c|}{\textbf{Configuration}}
  & \multicolumn{5}{c|}{\textbf{Back-translation}}
  & \multicolumn{3}{c}{\textbf{Pose quality}} \\
\cmidrule(lr){2-5}\cmidrule(lr){6-10}\cmidrule(lr){11-13}
  & MP & $N_q$ & Dilation & Length
  & WER$\downarrow$ & BLEU-1$\uparrow$ & BLEU-4$\uparrow$ & CHRF$\uparrow$ & ROUGE$\uparrow$
  & FID$\downarrow$ & MPJPE$\downarrow$ & MPJAE$\downarrow$ \\
\midrule
\textbf{SignRR (proposed)}
  & \cmark & 3 & mixed & Pred.
  & 80.98 & 28.78 & 10.96 & 32.90 & 28.15
  & 1.21 & 53.70 & 18.31 \\
\midrule
w/o MP
  & \xmark & 3 & mixed & Pred.
  & 82.25 & 27.89 & 10.70 & 32.32 & 26.94
  & 1.25 & 53.52 & 18.36 \\
\midrule
$N_q=1$
  & \cmark & 1 & mixed & Pred.
  & 81.52 & 26.82 & 10.57 & 31.82 & 26.58
  & 1.24 & 53.75 & 18.16 \\
$N_q=2$
  & \cmark & 2 & mixed & Pred.
  & 80.86 & 28.01 & 11.23 & 32.81 & 28.08
  & 1.23 & 53.48 & 18.29 \\
$N_q=4$
  & \cmark & 4 & mixed & Pred.
  & 80.32 & 27.43 & 10.61 & 32.12 & 27.46
  & 1.22 & 53.61 & 18.22 \\
\midrule
uniform dilation
  & \cmark & 3 & $[1,3,9]$ & Pred.
  & 80.70 & 27.63 & 10.83 & 32.10 & 27.13
  & 1.21 & 53.26 & 18.29 \\
uniform dilation
  & \cmark & 3 & $[1,3,5]$ & Pred.
  & 81.10 & 27.90 & 10.99 & 32.87 & 27.85
  & 1.21 & 53.26 & 18.29 \\
\midrule
w/o length prediction
  & \cmark & 3 & mixed & GT
  & 81.31 & 27.07 & 10.28 & 31.77 & 26.68
  & 1.16 & 47.61 & 17.81 \\
\bottomrule
\end{tabular}%
}
\end{table*}

\subsection{Analysis}

\smallskip\noindent\textbf{Disentangling Selection and Refinement.} SignRR combines a segment selector with a learned refinement model. To isolate their roles, we vary both parts on PHOENIX14T (Table~\ref{tab:analysis_selection_refinement}). For selection, we compare dynamic programming (DP) with random sampling during evaluation; training is unchanged and always uses randomly sampled segments. For refinement, we either apply the learned refiner or remove it, in which case the selected segments are directly concatenated and evaluated.

Refinement provides the main gain in both selection settings. With DP selection, it raises BLEU-1 from $22.06$ to $28.78$ and reduces MPJPE from $68.25$ to $53.70$. With random selection, it also raises BLEU-1 from $23.94$ to $28.23$. By contrast, DP selection alone does not improve over random selection in this setting, giving lower BLEU-1 without refinement ($22.06$ and $23.94$). Thus, the selector is not the main source of the improvement; its benefit appears only when the selected sequence is further refined.

With refinement enabled, DP selection gives a smaller but consistent gain over random selection, improving BLEU-1 by $+0.55$, BLEU-4 by $+0.20$, and ROUGE by $+0.51$. Refinement also makes the result less sensitive to the retrieved assembly: across three random-selection seeds, the standard deviation drops from $0.74$ to $0.06$ for BLEU-1, from $0.73$ to $0.25$ for ROUGE, and from $0.43$ to $0.03$ for MPJPE. These results show that refinement drives most of the improvement, while DP selection adds a small but stable gain when combined with the refiner.

\begin{table*}[!t]
\centering
\caption{Selection and refinement analysis of SignRR on PHOENIX14T (\textsc{Test}).  Selection denotes the segment retrieval policy used during evaluation, and Ref. indicates whether the part-aware Residual VQ-VAE is applied.  Rows with random selection are averaged over three seeds and reported as mean$\pm$std.}
\label{tab:analysis_selection_refinement}
\resizebox{\textwidth}{!}{%
\begin{tabular}{l|cc|ccccc|ccc}
\toprule
\multirow{2}{*}{\textbf{Variant}}
  & \multicolumn{2}{c|}{\textbf{Configuration}}
  & \multicolumn{5}{c|}{\textbf{Back-translation}}
  & \multicolumn{3}{c}{\textbf{Pose quality}} \\
\cmidrule(lr){2-3}\cmidrule(lr){4-8}\cmidrule(lr){9-11}
  & Selection & Refine.
  & WER$\downarrow$ & BLEU-1$\uparrow$ & BLEU-4$\uparrow$ & CHRF$\uparrow$ & ROUGE$\uparrow$
  & FID$\downarrow$ & MPJPE$\downarrow$ & MPJAE$\downarrow$ \\
\midrule
Retrieve only             & DP     & \xmark & 83.78          & 22.06          & 8.31           & 27.86          & 21.47          & 1.44          & 68.25          & 19.06 \\
Retrieve only             & Random & \xmark & 81.50$\pm$0.65 & 23.94$\pm$0.74 & 9.13$\pm$0.40  & 29.36$\pm$0.22 & 23.27$\pm$0.73 & 1.38$\pm$0.01 & 69.99$\pm$0.43 & 19.61$\pm$0.06 \\
SignRR (Random sel.)         & Random & \cmark & 81.11$\pm$0.21 & 28.23$\pm$0.06 & 10.76$\pm$0.17 & 32.61$\pm$0.14 & 27.64$\pm$0.25 & 1.21$\pm$0.01 & 53.74$\pm$0.03 & 18.32$\pm$0.01 \\
\textbf{SignRR (Proposed)} & DP    & \cmark & \textbf{80.98} & \textbf{28.78} & \textbf{10.96} & \textbf{32.90} & \textbf{28.15} & \textbf{1.21} & \textbf{53.70} & \textbf{18.31} \\
\bottomrule
\end{tabular}%
}
\end{table*}

\begin{table*}[!t]
\centering
\caption{Refinement without retrieval on PHOENIX14T (\textsc{Test}). 
\emph{Training} denotes whether the refiner is trained on real dictionary segments or synthetic periodic motion. 
\emph{Cross-attn} indicates whether the decoder attends to gloss features during refinement. 
\textbf{Bold} marks the best value in each column.}
\label{tab:analysis_refine_without_retrieval}
\resizebox{\textwidth}{!}{%
\begin{tabular}{l|cc|ccccc|ccc}
\toprule
\multirow{2}{*}{\textbf{Variant}}
  & \multicolumn{2}{c|}{\textbf{Configuration}}
  & \multicolumn{5}{c|}{\textbf{Back-translation}}
  & \multicolumn{3}{c}{\textbf{Pose quality}} \\
\cmidrule(lr){2-3}\cmidrule(lr){4-8}\cmidrule(lr){9-11}
  & Training & Cross-attn
  & WER$\downarrow$ & BLEU-1$\uparrow$ & BLEU-4$\uparrow$ & CHRF$\uparrow$ & ROUGE$\uparrow$
  & FID$\downarrow$ & MPJPE$\downarrow$ & MPJAE$\downarrow$ \\
\midrule
Synthetic input
  & Synthetic  & \xmark
  & 99.04 & 10.39 & 1.01  & 19.70 & 8.11
  & 1.87 & 56.19 & 19.77 \\
Synthetic input
  & Synthetic  & \cmark
  & 90.02 & 25.06 & 8.72  & 30.13 & 23.99
  & 1.36 & \textbf{49.96} & 18.60 \\
\textbf{SignRR (Proposed)}
  & Dictionary & \xmark
  & \textbf{80.98} & \textbf{28.78} & \textbf{10.96} & \textbf{32.90} & \textbf{28.15}
  & \textbf{1.21} & 53.70 & 18.31 \\
SignRR + gloss
  & Dictionary & \cmark
  & 83.47 & 27.03 & 10.24 & 31.84 & 27.17
  & 1.24 & 53.64 & \textbf{18.08} \\
\bottomrule
\end{tabular}%
}
\end{table*}

\smallskip\noindent\textbf{Refinement without Retrieval.} The previous analysis shows that refinement is an important part of the retrieve-and-refine pipeline, but it still uses real retrieved segments as input. This raises a key question: can refinement work equally well without real retrieved motion? We address this by comparing refiners trained on either real dictionary segments or synthetic periodic motion (Table~\ref{tab:analysis_refine_without_retrieval}). The synthetic motion is generated from periodic trajectories, so it provides smooth pose variation but no real sign articulation. We also add an optional decoder cross-attention layer, where the decoder attends to gloss features during refinement, to test whether explicit gloss conditioning can compensate for the missing retrieved motion.

Synthetic periodic motion performs poorly without decoder cross-attention, reaching WER $99.04$ and BLEU-1 $10.39$. Adding cross-attention substantially improves this setting, increasing BLEU-1 to $25.06$ and ROUGE to $23.99$. However, it still remains below SignRR trained on real dictionary segments, which reaches BLEU-1 $28.78$, BLEU-4 $10.96$, and ROUGE $28.15$. Adding cross-attention to real retrieved motion also brings no consistent benefit, lowering test BLEU-1 from $28.78$ to $27.03$ and increasing WER from $80.98$ to $83.47$. These results indicate that gloss conditioning alone cannot replace real retrieved motion; the dictionary segments provide sign-specific articulation and dynamics that synthetic motion does not capture.

Overall, these two analyses support our retrieve-and-refine design. The first shows that refinement improves and stabilizes retrieved assemblies, while the second shows that refinement on real retrieved segments outperforms refinement on synthetic motion, even with explicit gloss cross-attention. Retrieval provides sign-specific motion cues that the synthetic input cannot fully provide, and refinement converts them into a more coherent signing sequence.

\smallskip\noindent\textbf{Robustness to Segmentation Error.} Since the dictionary is built from CorrNet+ boundaries, we analyze how segmentation error affects production quality by perturbing gloss boundaries by $\pm4$, $\pm8$, and $\pm16$ frames, and by replacing the recognizer with uniform segmentation, rebuilding the dictionary and retraining the model in each case. BLEU-4 decreases by at most $0.42$ at $\pm16$ frames, while the uniform-segmentation setting still reaches BLEU-4 $10.58$ on the test set. The full experiment and results are provided in the supplementary material.

\smallskip\noindent\textbf{Inference Efficiency.} SignRR requires $87.8$\,ms per sequence on the PHOENIX14T test set with batch size $1$ on a single A100 GPU. This is about $25\times$ faster than G2P-DDM with $100$ diffusion steps and has similar latency to Sign-IDD with $10$ DDIM steps. The full experiment and protocol information are provided in the supplementary material.

\section{Limitations}
\label{sec:limitations}

Currently, SignRR focuses on body and hand articulation. The 61-keypoint representation includes a set of facial landmarks, but it does not capture fine-grained non-manual cues such as mouthings or eye movements, which can carry grammatical and lexical information in signed languages~\cite{sandler2006sign}. Therefore, adding a dedicated facial stream to the part-aware representation is an important direction for future work. Also, the retrieval stage depends on the sign inventory available in the gloss-motion dictionary. When a gloss is not present, our current implementation maps it to the closest dictionary entry using edit distance. A more general open-vocabulary setting could instead use external lexical resources or a generative
fallback for unseen signs. Finally, our evaluation follows the standard back-translation protocol, which assesses production quality indirectly through recognition performance. Future work could complement these metrics with evaluations by fluent signers to provide a more direct assessment of perceptual naturalness, articulation quality, and linguistic intelligibility.

\section{Conclusion}

We presented SignRR, a retrieve-and-refine framework for sign language production. Rather than generating signing motion from scratch, SignRR retrieves real sign segments from a gloss-motion dictionary, assembles them into an initial sequence, and refines the full motion with a part-aware Residual VQ-VAE. This design uses retrieval to preserve realistic articulation and learned refinement to improve coherence. Experiments on PHOENIX14T and CSL-Daily show that SignRR achieves state-of-the-art back-translation performance among non-GT methods while maintaining competitive pose quality. Ablation studies further show that full-sequence refinement is important to the method and improves the quality of the retrieved assemblies. Future work will connect the refined pose sequences with photorealistic avatars for end-to-end sign video synthesis.

\vspace{1em}
\noindent\textbf{Acknowledgment.} This work was funded by the Schmidt Sciences AI2050 Early Career Fellowship, Grant G-23-66054.

\bibliography{egbib}

%-------------------------------------------------------------------------
% Supplementary material starts here
%-------------------------------------------------------------------------
\clearpage

% Reset counters and re-letter sections/figures/tables/equations as A, A.1, ...
\setcounter{section}{0}
\setcounter{subsection}{0}
\setcounter{figure}{0}
\setcounter{table}{0}
\setcounter{equation}{0}
\renewcommand{\thesection}{\Alph{section}}
\renewcommand{\thesubsection}{\Alph{section}.\arabic{subsection}}
\renewcommand{\thefigure}{A\arabic{figure}}
\renewcommand{\thetable}{A\arabic{table}}
\renewcommand{\theequation}{A\arabic{equation}}

\begin{center}
{\Large\bfseries Supplementary Material}
\end{center}
\bigskip

\section{Keypoint Representation}
\label{sec:supp_keypoints}

\begin{figure}[h]
    \centering
    \includegraphics[width=0.8\linewidth]{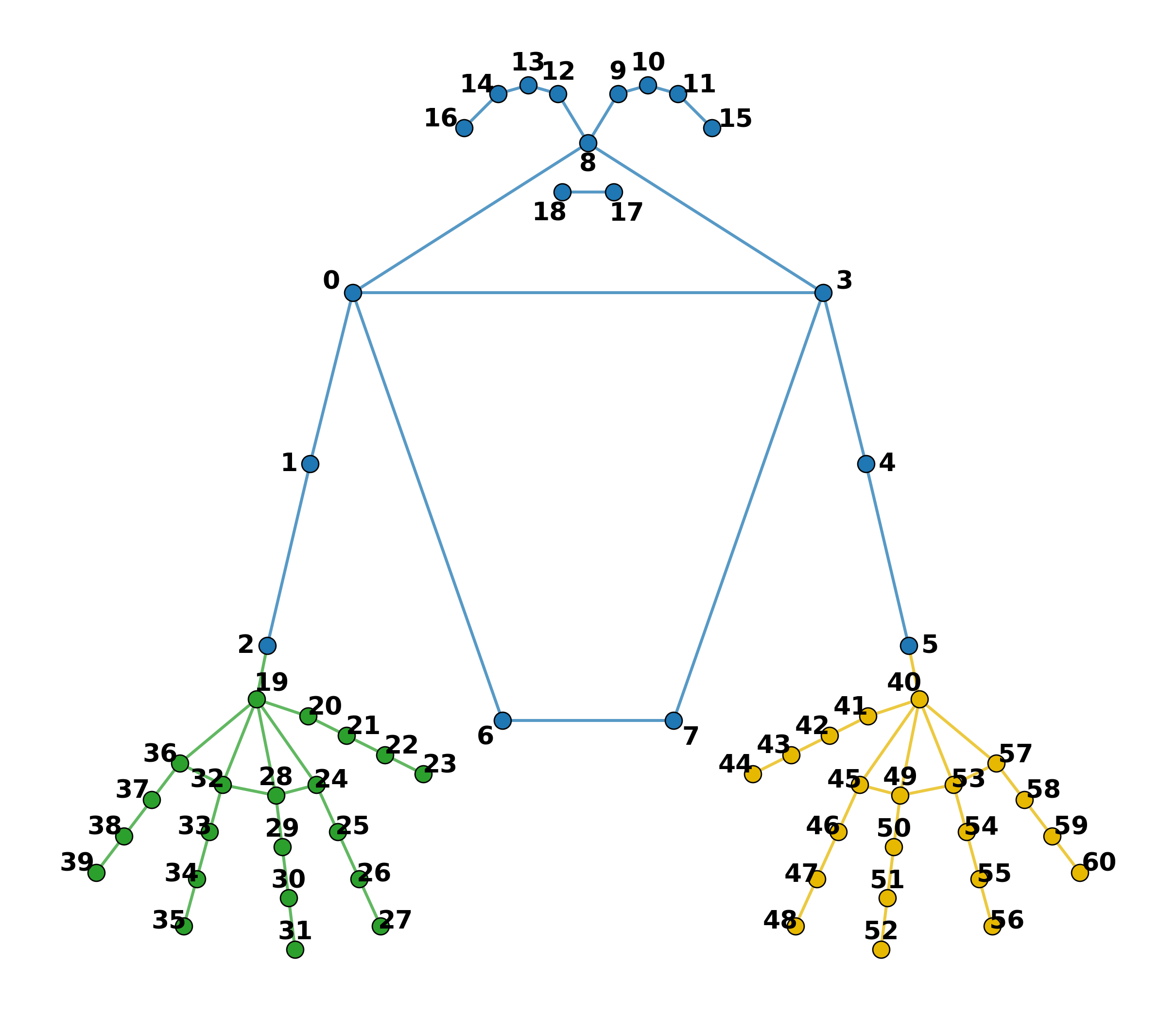}
    \caption{Keypoint layout. Blue nodes denote body keypoints, green nodes denote right-hand keypoints, and yellow nodes denote left-hand keypoints.}
    \label{fig:keypoint_layout}
\end{figure}

We use a 61-keypoint 3D pose representation for each frame. Each keypoint contains $(x,y,z)$ coordinates, giving $D=61 \times 3 = 183$ pose values per frame. The layout is shown in Fig.~\ref{fig:keypoint_layout}.

\begin{itemize}
    \item \textbf{Body keypoints (0--18):} upper torso, arms, hips, and facial landmarks. This includes shoulders, elbows, wrists, hips, the neck/head base, and facial points around the eyes, ears, and mouth.

    \item \textbf{Right-hand keypoints (19--39):} one wrist keypoint and five four-joint finger chains.

    \item \textbf{Left-hand keypoints (40--60):} one wrist keypoint and five four-joint finger chains, following the same layout as the right hand.
\end{itemize}

\section{Gloss-Motion Dictionary Construction}
\label{sec:supp_dictionary}

We construct the gloss-motion dictionary from the training split. The dictionary stores multiple real motion segments for each gloss, which are later used to initialize the retrieved sequence in SignRR. Since the datasets provide sentence-level gloss annotations but not frame-level gloss boundaries, we estimate segment boundaries using CorrNet+~\cite{hu2024corrnet+} and the known gloss order.

\paragraph{Boundary estimation.}
Given a training sequence with glosses $G=(g_1,\dots,g_N)$, we run the pretrained CorrNet+ recognizer to obtain a frame-level score $r_t(g)$ for each gloss $g$ at frame $t$. We then use the ground-truth gloss order to estimate the start of each gloss. For gloss $g_i$, the start frame is selected as the first frame after the previous boundary where the CorrNet+ score of $g_i$ becomes larger than the score of the previous gloss:
\begin{equation}
b_i = \min \{t > b_{i-1} \mid r_t(g_i) > r_t(g_{i-1}) \}.
\label{eq:supp_boundary}
\end{equation}
This gives a weak temporal alignment without requiring frame-level boundary annotations. The segment for $g_i$ is extracted from the interval between consecutive boundaries.

\paragraph{Candidate filtering.}
The raw aligned segments can contain noisy boundaries or abnormal durations. We therefore filter candidates independently for each gloss. Let $\ell(s)$ denote the duration of segment $s$, and let $\mu_g$ and $\sigma_g$ be the mean and standard deviation of the segment durations for gloss $g$. A segment is discarded if its duration is too far from the typical duration of that gloss:
\begin{equation}
|\ell(s)-\mu_g| > \tau_{\ell}\sigma_g .
\label{eq:supp_duration_filter}
\end{equation}
This removes very short or very long segments that are likely to come from poor alignment.

\paragraph{Candidate scoring.}
After filtering, each remaining segment is scored using motion smoothness and duration regularity. The smoothness term favors segments with stable temporal motion. For a segment $s=(\mathbf{x}_1,\dots,\mathbf{x}_{\ell})$, we measure smoothness using the average frame-to-frame acceleration:
\begin{equation}
e_{\mathrm{sm}}(s)
=
\frac{1}{\ell-2}
\sum_{t=2}^{\ell-1}
\left\|
\mathbf{x}_{t+1}-2\mathbf{x}_{t}+\mathbf{x}_{t-1}
\right\|_2 .
\label{eq:supp_smoothness_error}
\end{equation}
Lower values indicate smoother motion. We convert this error into a normalized score:
\begin{equation}
q_{\mathrm{sm}}(s) = \exp\left(-e_{\mathrm{sm}}(s)\right).
\label{eq:supp_smoothness_score}
\end{equation}

The duration term favors segments whose length is close to the typical duration of the same gloss:
\begin{equation}
q_{\mathrm{dur}}(s)
=
\exp\left(
-\frac{|\ell(s)-\mu_g|}{\sigma_g+\epsilon}
\right).
\label{eq:supp_duration_score}
\end{equation}
The final candidate score combines both terms:
\begin{equation}
q(s)
=
\lambda_{\mathrm{sm}} q_{\mathrm{sm}}(s)
+
\lambda_{\mathrm{dur}} q_{\mathrm{dur}}(s),
\label{eq:supp_candidate_score}
\end{equation}
where $\lambda_{\mathrm{sm}}$ and $\lambda_{\mathrm{dur}}$ control the relative weights of motion smoothness and duration regularity. For each gloss, we keep the top $S$ candidates according to $q(s)$.

\paragraph{Boundary descriptors.}
For each retained segment, we also store a boundary descriptor used by the dynamic-programming selector during inference. The descriptor contains the first and last poses of the segment, together with their local velocities:
\begin{equation}
\phi(s)
=
\left[
\mathbf{x}_1,\ 
\mathbf{x}_{\ell},\ 
\mathbf{x}_2-\mathbf{x}_1,\ 
\mathbf{x}_{\ell}-\mathbf{x}_{\ell-1}
\right].
\label{eq:supp_boundary_descriptor}
\end{equation}
This descriptor allows the selector to compare the compatibility of adjacent retrieved segments without using the full segment sequence.

\begin{algorithm}[t]
\caption{Gloss-motion dictionary construction}
\label{alg:supp_dictionary}
\begin{algorithmic}[1]
\Require Training sequences with sentence-level gloss annotations; pretrained CorrNet+ recognizer
\Ensure Gloss-motion dictionary $\mathcal{D}$

\State Initialize an empty candidate set for each gloss

\For{each training sequence}
    \State Run CorrNet+ to obtain frame-level gloss scores
    \State Estimate gloss boundaries using Eq.~\ref{eq:supp_boundary}
    \For{each aligned gloss occurrence}
        \State Extract the corresponding motion segment
        \State Add the segment to the candidate set of that gloss
    \EndFor
\EndFor

\For{each gloss $g$}
    \State Compute duration statistics $\mu_g$ and $\sigma_g$
    \State Remove abnormal-duration segments using Eq.~\ref{eq:supp_duration_filter}
    \State Score each remaining segment using Eq.~\ref{eq:supp_candidate_score}
    \State Keep the top $S$ candidates
    \State Store each retained segment and its boundary descriptor in $\mathcal{D}(g)$
\EndFor

\State \Return $\mathcal{D}$
\end{algorithmic}
\end{algorithm}

The resulting dictionary provides real motion candidates for each gloss. The filtering step reduces noisy alignments, while the scoring step favors segments that are both temporally smooth and close to the typical duration of the corresponding gloss.

\section{Dynamic Programming Segment Selection}
\label{sec:supp_dp}

At inference time, SignRR receives a gloss sequence and selects one motion segment per gloss from the dictionary $\mathcal{D}$. Selecting segments only by their individual quality can produce visible jumps at segment boundaries, while selecting only by boundary smoothness can ignore the quality of the retrieved signs. We therefore optimize both terms jointly using dynamic programming over the candidate trellis defined by $\mathcal{D}$.

\paragraph{Out-of-dictionary glosses.}
A gloss in the test input may be absent from $\mathcal{D}$ because it was filtered out during dictionary construction (Sec.~\ref{sec:supp_dictionary}) or because of minor spelling variants. We replace each such gloss with the closest dictionary entry $g^\star$ under Levenshtein edit distance:
\begin{equation}
g^\star = \argmin_{g' \in \mathcal{V}_{\mathcal{D}}}\; \mathrm{Lev}(g, g'),
\label{eq:supp_oov}
\end{equation}
where $\mathcal{V}_{\mathcal{D}}$ is the set of glosses present in the dictionary. This ensures that each gloss position has at least one candidate segment. In the two evaluated datasets, unseen gloss types do not often occur. PHOENIX14T contains $19$ in the development split and $22$ in the test split, while CSL-Daily contains none.

\paragraph{Boundary cost.}
The boundary cost measures the discontinuity between two consecutive segments. Given a previous segment $s'$ and a current segment $s$, we compare the end descriptor of $s'$ with the start descriptor of $s$:
\begin{equation}
d_{\mathrm{bnd}}(s',s)
=
\big\|\, \phi_{\mathrm{end}}(s') - \phi_{\mathrm{start}}(s) \,\big\|_2 .
\label{eq:supp_boundary_cost}
\end{equation}
The descriptors include both pose and local velocity, so this cost penalizes spatial jumps and abrupt motion changes at the seam. Since $d_{\mathrm{bnd}}$ and the segment quality score $q(s) \in [0,1]$ have different scales, we normalize the boundary cost by a dataset-level constant $\tilde{d}$. We compute $\tilde{d}$ once as the median boundary cost over randomly sampled pairs of dictionary segments, making the normalization robust to outliers.

\paragraph{Dynamic-programming recurrence.}
Let $\mathcal{S}_i=\mathcal{D}(g_i)$ be the candidate set for gloss $g_i$, and let $s_{i,j}\in\mathcal{S}_i$ denote its $j$-th candidate. We define $C_i(j)$ as the minimum cumulative cost of any path ending at $s_{i,j}$. The first position is initialized using only the segment quality cost:
\begin{equation}
C_1(j) = \beta\big(1-q(s_{1,j})\big).
\label{eq:supp_dp_init}
\end{equation}
For $i=2,\dots,N$, the recurrence is
\begin{equation}
C_i(j)
=
\min_k
\left[
C_{i-1}(k)
+
\alpha \frac{d_{\mathrm{bnd}}(s_{i-1,k},s_{i,j})}{\tilde{d}}
\right]
+
\beta\big(1-q(s_{i,j})\big),
\label{eq:supp_dp_recurrence}
\end{equation}
where $\alpha$ and $\beta$ control the trade-off between transition smoothness and segment quality. We store the minimizing predecessor in a back-pointer table $B_i(j)$ and use $\alpha=1.0$ and $\beta=0.5$ in all experiments.

\paragraph{Backtracking.}
After the forward pass, we start from the lowest-cost candidate at the final gloss:
\begin{equation}
j^\star_N = \argmin_j C_N(j).
\label{eq:supp_dp_final}
\end{equation}
We then recover the full path by following the back-pointers:
\begin{equation}
j^\star_i = B_{i+1}(j^\star_{i+1}),
\qquad i=N-1,\dots,1.
\label{eq:supp_dp_backtrack}
\end{equation}
The selected path is $\pi^\star=(\pi^\star_1,\dots,\pi^\star_N)$, where $\pi^\star_i=s_{i,j^\star_i}$. These segments are concatenated in order and passed to the RVQ-VAE refinement model as the initial motion.

\begin{algorithm}[t]
\caption{Dynamic programming segment selection}
\label{alg:supp_dp}
\begin{algorithmic}[1]
\Require Gloss sequence $G=(g_1,\dots,g_N)$; dictionary $\mathcal{D}$; weights $\alpha,\beta$; normalizer $\tilde{d}$
\Ensure Selected path $\pi^\star=(\pi^\star_1,\dots,\pi^\star_N)$

\State Resolve out-of-dictionary glosses in $G$ using Eq.~\ref{eq:supp_oov}
\State Let $\mathcal{S}_i=\mathcal{D}(g_i)$ for $i=1,\dots,N$

\For{$j=1,\dots,|\mathcal{S}_1|$}
    \State $C_1(j) \gets \beta\big(1-q(s_{1,j})\big)$
\EndFor

\For{$i=2,\dots,N$}
    \For{$j=1,\dots,|\mathcal{S}_i|$}
        \State Compute $C_i(j)$ using Eq.~\ref{eq:supp_dp_recurrence}
        \State Store the minimizing predecessor in $B_i(j)$
    \EndFor
\EndFor

\State $j^\star_N \gets \argmin_j C_N(j)$
\For{$i=N-1,\dots,1$}
    \State $j^\star_i \gets B_{i+1}(j^\star_{i+1})$
\EndFor

\State $\pi^\star_i \gets s_{i,j^\star_i}$ for $i=1,\dots,N$
\State \Return $\pi^\star$
\end{algorithmic}
\end{algorithm}

With at most $S$ candidates per gloss and a gloss sequence of length $N$, the algorithm runs in $O(NS^2)$ time. The selected path balances segment quality and boundary continuity, giving the refinement model a smoother retrieved initialization.

\section{Implementation Details}
\label{sec:supp_impl}

This section summarizes the implementation details used in our experiments.

\paragraph{Refinement model.}
The refinement model uses part-specific encoders for the body, right hand, and left hand. Each encoder has two stride-2 downsampling stages with dilated ResNet-1D blocks and hidden width $512$. Each part uses $L=3$ residual codebooks. The body stream uses codebooks with $256$ entries, while each hand stream uses codebooks with $384$ entries. The shared decoder mirrors the encoder structure and outputs the $D=183$ pose dimensions.

\paragraph{Gloss encoder and length prediction.}
The gloss encoder is a $2$-layer Transformer with $4$ attention heads, model dimension $128$, and dropout $0.1$. Its mean-pooled output is projected to $512$ dimensions and passed to a $2$-layer ReLU MLP with hidden width $256$ to predict the length offset $\Delta$. We use $T_{\max}=300$ for PHOENIX14T, $T_{\max}=352$ for CSL-Daily, and $\delta_{\mathrm{scale}}=100$. At inference, the predicted target length is clamped to $[16,T_{\max}]$.

\paragraph{Training objective.}
For the training objective in the main paper, we use $\lambda_v=0.1$, $\lambda_c=0.02$, and $\lambda_\ell=0.01$. In the part-weighted smooth-$\ell_1$ reconstruction loss, the body weight is $1.0$, and each hand weight is $2.0$.

\paragraph{Dictionary configuration.}
For each gloss, we keep at most $S=25$ representative segments. Candidates are selected using the quality score described in Sec.~\ref{sec:supp_dictionary}, with diversity-based subsampling when more than $S$ candidates are available. The dynamic-programming normalizer $\tilde{d}$ is estimated once per dataset from randomly sampled inter-gloss segment pairs. After filtering, the dictionaries contain $9{,}311$ segments for PHOENIX14T and $32{,}500$ segments for CSL-Daily, with an average of $8.61$ and $16.26$ candidates per gloss, respectively.

\section{Synthetic Motion Generation}
\label{sec:supp_synthetic}

For our refinement-without-retrieval analysis, we use a synthetic motion to evaluate whether the refinement model can operate without real retrieved sign segments. We synthesized continuous hand displacement trajectories by adding mathematically defined motion curves to a randomly selected reference frame. We used three trajectory types to approximate simple hand motion patterns: linear (Eq.~\ref{eq:supp_linear_case}), sinusoidal (Eq.~\ref{eq:supp_sinusoidal_case}), and quadratic (Eq.~\ref{eq:supp_quadratic_case}). Each trajectory used a fixed amplitude chosen empirically, while the orientation and frequency were randomized to increase variation across samples.

Let $n$ denote the number of frames in the synthetic sequence, and let $s \sim \mathcal{U}(0.005, 0.02)$ be the step size sampled uniformly at random. The displacement domain is defined as an evenly spaced sequence of $n$ points. Each trajectory returns a sequence of 2D displacement vectors $\{(x_i, y_i)\}_{i=0}^{n-1}$, where $x_i = i \cdot s$ and $i = 0, 1, \dots, n-1$. The three trajectories are defined as follows:

\begin{equation}
(x_i,\ y_i) = \begin{cases}
(0,\ i \cdot s) & \text{if vertical}, \\
(i \cdot s,\ 0) & \text{if horizontal}.
\end{cases}
\label{eq:supp_linear_case}
\end{equation}

\begin{equation}
x_i = i \cdot s, \qquad y_i = 12\, x_i^2
\label{eq:supp_quadratic_case}
\end{equation}

\begin{equation}
x_i = i \cdot s, \qquad y_i = 0.1 \sin(25\, x_i)
\label{eq:supp_sinusoidal_case}
\end{equation}

The resulting displacement vectors are added to the X and Y components of the keypoints from the randomly selected reference frame $\mathbf{p}^{(\text{ref})}$, producing the displaced keypoint positions $\mathbf{p}_i$:

\begin{equation}
\mathbf{p}_i = \mathbf{p}^{(\text{ref})} + \mathbf{d}_i,
\quad \mathbf{d}_i = (x_i,\ y_i) \in \mathbb{R}^2 .
\end{equation}

The trajectories were applied to 46 selected keypoints per frame, since these keypoints carry most of the movement during signing. The selected keypoints cover the wrists, palms, and finger joints of both hands. Specifically, the right hand includes keypoints 1--2 and 19--39, while the left hand includes keypoints 4--5 and 40--60.

To produce temporally smooth sequences, adjacent segments were joined using linear boundary interpolation (Eq.~\ref{eq:supp_interp_equ}). Given two consecutive segments $S^{(k)}$ and $S^{(k+1)}$, let $\mathbf{p}^{(k)}_{\text{last}}$ and $\mathbf{p}^{(k+1)}_{\text{first}}$ denote their boundary frames. A fixed number of $T=10$ interpolated frames is inserted between them:

\begin{equation}
\mathbf{p}_t = (1 - \alpha_t)\, \mathbf{p}^{(k)}_{\text{last}}
+ \alpha_t\, \mathbf{p}^{(k+1)}_{\text{first}},
\quad t = 1, \dots, T,
\qquad \alpha_t = \frac{t}{T + 1}.
\label{eq:supp_interp_equ}
\end{equation}

Since $\alpha_t \in (0, 1)$, the boundary frames are not duplicated, and the transitions remain smooth. All segments are then concatenated to produce synthetic sequences with lengths comparable to the original sequences, allowing a variation of $\pm 20$ frames per video.

We then applied a filtering step to remove physically implausible frames based on two anatomical constraints. A frame is marked as invalid if either of the following rules is violated:

\begin{itemize}
\item \textbf{Hand height:} no keypoint belonging to either hand may exceed a vertical ceiling defined by keypoint 9 plus an offset of 0.6 units. This prevents the hands from being placed far above the head.
\item \textbf{Arm length:} the elbow keypoints, keypoints 1 and 5, must remain within 20\% of the vertical distance between keypoints 0 and 6, which corresponds to the torso length. This prevents physically implausible arm extensions.
\end{itemize}

After filtering, the remaining valid frames are resampled back to the original sequence length using linear interpolation, following the same procedure as Eq.~\ref{eq:supp_interp_equ}.

The resulting synthetic sequences approximate simple continuous hand motion patterns while remaining semantically neutral; that is, they do not correspond to any actual sign or meaningful gesture. This provides smooth motion inputs that lack sign-specific articulation, allowing us to test whether real retrieved motion is necessary for effective refinement.

\section{Robustness to Dictionary Segmentation}
\label{sec:supp_robust}

We evaluate the robustness of SignRR to variations in the gloss boundaries used to construct the dictionary. For each interior boundary, we independently sample an offset from the discrete uniform distribution over $\{-k,\dots,k\}$ frames, with $k \in \{4,8,16\}$, and shift the boundary by that offset. We constrain the perturbed boundaries to preserve gloss order and a minimum segment length of three frames, while keeping the first and last sequence boundaries fixed. For each setting, we rebuild the dictionary from the perturbed segments and retrain SignRR using the same configuration as in the main experiments.

We also consider a recognizer-free baseline in which a sequence of length $L$ containing $N$ glosses is uniformly divided into $N$ segments without using CorrNet+ predictions. The same dictionary construction and training procedure is then applied. Table~\ref{tab:supp_robust} shows that performance remains stable under perturbations and decreases gradually as the boundary variation increases. At $\pm16$ frames, BLEU-4 decreases by only $0.42$. Uniform segmentation still obtains BLEU-4 $10.58$, showing that CorrNet+-based boundaries are helpful, while SignRR remains effective with a much coarser segmentation.

\begin{table}[t]
\centering
\caption{Robustness to dictionary segmentation on PHOENIX14T (\textsc{Test}). Each setting
rebuilds the dictionary and retrains SignRR using the corresponding boundaries.}
\label{tab:supp_robust}
\footnotesize
\setlength{\tabcolsep}{6pt}
\begin{tabular}{l|ccccc}
\toprule
\textbf{Dictionary} & WER$\downarrow$ & BLEU-1$\uparrow$ & BLEU-4$\uparrow$ & ROUGE$\uparrow$ & MPJPE$\downarrow$ \\
\midrule
SignRR (proposed)
& 80.98 & 28.78 & 10.96 & 28.15 & 53.70 \\
Jitter $\pm4$
& 81.43 & 28.28 & 11.28 & 27.90 & 53.47 \\
Jitter $\pm8$
& 82.08 & 27.57 & 10.65 & 27.21 & 54.05 \\
Jitter $\pm16$
& 82.88 & 27.00 & 10.54 & 26.53 & 54.41 \\
Uniform (no recognizer)
& 81.40 & 27.26 & 10.58 & 27.12 & 53.55 \\
\bottomrule
\end{tabular}
\end{table}

\begin{table}[t]
\centering
\caption{Per-sequence inference time on PHOENIX14T (\textsc{Test}) with batch size $1$ on a
single NVIDIA A100 GPU.}
\label{tab:supp_runtime}
\footnotesize
\setlength{\tabcolsep}{6pt}
\begin{tabular}{lrr}
\toprule
\textbf{Method} & \textbf{ms/seq.} & \textbf{Params (M)} \\
\midrule
SignRR: DP selection   & 54.6  & -- \\
SignRR: assembly       & 0.1   & -- \\
SignRR: refinement     & 33.1  & 38.5 \\
\textbf{SignRR: total} & \textbf{87.8} & \textbf{38.5} \\
\midrule
Sign-IDD~\cite{tang2025sign} & 85.0  & 16.8 \\
G2P-DDM~\cite{xie2024g2p}    & 2237  & 106.3 \\
\bottomrule
\end{tabular}
\end{table}

\section{Inference Time}
\label{sec:supp_runtime}

We evaluate inference time on the PHOENIX14T test set with batch size $1$ on a single NVIDIA A100 GPU. Dynamic-programming selection runs on the CPU, while refinement is measured as one forward pass of the full model.

Table~\ref{tab:supp_runtime} reports the inference time for each component and the parameter counts. Sign-IDD and G2P-DDM are measured using their released implementations on the same GPU with $10$ DDIM and $100$ diffusion steps, respectively. SignRR requires $87.8$\,ms per sequence, with $54.6$\,ms for segment selection and $33.1$\,ms for refinement. Its total inference time is comparable to Sign-IDD and about $25\times$ faster than G2P-DDM.

\end{document}